# A Two-Stage Combined Classifier in Scale Space Texture Classification

Mehrdad J. Gangeh[a,∗], Robert P.W. Duin[b], Bart M. ter Haar Romeny[c], Mohamed S. Kamel[a]

[a]*Center for Pattern Analysis and Machine Intelligence, Department of Electrical and Computer Engineering, University of Waterloo, 200 University Avenue West, Waterloo, ONT. N2L 3G1, Canada*
[b]*Pattern Recognition Laboratory, Delft University of Technology, Mekelweg 4, 2628 CD Delft, The Netherlands*
[c]*Biomedical Image Analysis Group, Department of Biomedical Engineering, Eindhoven University of Technology, Den Dolech 2, NL-5600 MB Eindhoven, the Netherlands*

**Abstract**

Textures often show multiscale properties and hence multiscale techniques are considered useful for texture analysis. Scale-space theory as a biologically motivated approach may be used to construct multiscale textures. In this paper various ways are studied to combine features on different scales for texture classification of small image patches. We use the N-jet of derivatives up to the second order at different scales to generate distinct pattern representations (DPR) of feature subsets. Each feature subset in the DPR is given to a base classifier (BC) of a two-stage combined classifier. The decisions made by these BCs are combined in two stages over scales and derivatives. Various combining systems and their significances and differences are discussed. The learning curves are used to evaluate the performances. We found for small sample sizes combining classifiers performs significantly better than combining feature spaces (CFS). It is also shown that combining classifiers performs better than the support vector machine on CFS in multiscale texture classification.

## 1. Introduction

There is a vast literature on texture analysis, as can be judged from its numerous applications in various fields [1]. Texture analysis has been applied to nine broad categories [1–7]:

∗Corresponding author
*Email addresses:* `mgangeh@pami.uwaterloo.ca` (Mehrdad J. Gangeh), `r.duin@ieee.org` (Robert P.W. Duin), `B.M.terHaarRomeny@tue.nl` (Bart M. ter Haar Romeny), `mkamel@pami.uwaterloo.ca` (Mohamed S. Kamel)

1. *Texture classification* of stationary texture images that contain only one texture type per image.
2. Unsupervised *texture segmentation* of nonstationary images that consist of more than one texture type per image.
3. *Texture synthesis*, which is important in computer graphics for rendering object surfaces to be as realistic looking as possible.
4. *3D shape from texture*, which investigates how a standard texel shape is distorted by 3D projections and relates it to the local surface orientation.
5. *Shape from texture*, in which texture is usually used in addition to other features such as shading, color, and so on to extract three-dimensional shape information.
6. *Color-texture analysis*, where joint color-texture descriptor improves discrimination over using color and texture features independently.
7. *Texture for appearance modeling*, which is fundamental in computer vision and graphics.
8. *Dynamic texture analysis* for dynamic shape and appearance modeling which is essential in video sequences with certain temporal regularity properties.
9. *Indexing* and image database retrieval usually based on similarity measure.

In this paper, the focus is on texture classification among the texture analysis problems and hence in the remaining of the text, all the examples and statements are for this problem.

As texture is a complicated phenomenon, there is no unique definition that is agreed upon by the researchers [8]. However, for the purpose of this paper, we adopt to the definition provided in [5]: "Texture is the variation of data at scales smaller than the scales of interest".

Textures often show multiscale/multiresolution properties. This means that information from other scales can help to perform a more accurate texture analysis or, for example in texture classification, enables to train the classifier(s) by a smaller training set. This inspires the use of multiresolution techniques in texture analysis.

*1.1. Studies on Multiresolution Texture Analysis*

Multiresolution techniques in texture analysis can be divided into two broad categories: techniques based on a single approach and those which are based on a combination of approaches.

Some of the most well-known multiresolution techniques on texture analysis in the former category given above are: multiresolution histograms [2, 9] including locally orderless images [10], techniques based on multiscale local autocorrelation features [11], multiresolution local binary patterns [12, 13], multiresolution Markov random fields [14–17], wavelets [4, 18–22], Gabor filters [19, 23, 24], multiscale tensor voting [25], a technique based on multiresolution fractal feature vectors [26], and techniques based on scale-space theory [10, 27–29]. In these approaches, feature subsets from different scales are fused to construct a single feature space. This space is then submitted to a single classifier. These approaches report better results than those based on a single resolution.

Two research works in the literature that report fusion of two or more multiresolution techniques are: fusion of dyadic wavelet transform, steerable pyramid, Gabor wavelet transform, and wavelet frame transform in [30], and combining Laws filters, Gabor, and wavelets in [31]. These papers also report better results than single multiscale techniques.

*1.2. Contributions*

The main issue in multiresolution techniques is the large feature space generated. The common trend in the literature, which is the fusion of the feature subsets generated at different scales to come up with one feature space to be submitted to a classifier, makes the problem even more serious. The large feature space, however, suffers from the 'curse of dimensionality' [32]. To tackle this problem typically severe feature reduction is applied in multiresolution techniques, e.g. by using histogram moments [10, 27, 28], calculating the energy [19], or maximum entropy [17]. However, the performance of the classifiers trained for such a reduced feature space highly depends on how well these features represent the data in the particular application. This problem is also addressed in the literature by using classifiers that behave better in high dimensional feature space, e.g., support vector machines (SVMs) [9, 24].

Also, the multiresolution papers in texture classification only report the classification error for a single specific (usually large) training set size. This keeps the behavior of the classifier unrevealed in small training set sizes that might be important in some applications especially those where obtaining a large training set size is difficult, costly or even impossible. This is particularly the case in texture classification applications on medical images: obtaining medical images for some specific diseases is cumbersome especially in the case that standardized protocols for image acquisition are difficult to be followed, such as in ultrasound images [33].

This paper addresses the following two issues. First, we use combined classifiers instead of fusing feature subsets. In combined classifiers, feature subsets produced at various scales are submitted to base classifiers and hence feature fusion is no longer required. Next, the outputs of these base classifiers are combined for a final decision on the class of each texture. We propose a two-stage structure for this purpose and discuss the benefits obtained. We also provide the mathematical formulation for the proposed two-stage combined classifiers and its relation to one-stage structure. Second, the learning curves for training the classifiers are constructed for different training set sizes. This clearly shows how the training of the classifier evolves as we increase the training set size. Our focus is on the classification of small patches, which is a more challenging task as the information contained is less than large ones.

Our results lead to an important conclusion: convolution with filters like wavelets, Gabor, and Gaussian derivatives, which are used for the construction of feature subsets at different scales/resolutions, consists of linear operations. If there are no zero-weights, no information is lost as they are invertible. All information is available in the original space, i.e., at highest resolution and can be used for classification provided that the set of training samples is sufficiently large. As this is usually not possible or prohibitive, the information from other scales is provided to train the classifier using fewer data samples.

*1.3. Previous Work on Combined Classifiers in Multiresolution Texture Analysis*

A few papers in the literature examine the application of multiple classifier systems (MCS) in multiresolution texture classification. The application of MCS in multiresolution texture classification using SVMs as *base classifiers* (BCs) and wavelets as the multiresolution technique is investigated in [34]. The SVMs use different kernels. Combining them by majority voting produced better results than a single SVM if applied to the Brodatz album. Also [35] addresses the classification of multispectral images with the application in remote sensing using ensemble of classifiers with SVMs as the base classifiers. The results show improvement over single SVM.

On the other hand, it is reported in [36] that using combined classifiers to combine three different multiresolution techniques, i.e., Gabor, wavelets and the combination of these two using $k$-NN or SVMs as base classifier does not improve over the best single classifier for scenery images.

However, to our best of knowledge, there is no work in the literature on using combined classifiers where the feature subsets from different scales are submitted to the BCs. In this paper, we will discuss this approach and the benefits obtained.

Scale space theory in the context of multiscale texture classification is presented in Section 2. In Section 3, combined classifiers and their application in scale space texture classification are explained. The experiments are elaborated in Section 4 followed by the results in Section 5. The comparison between the proposed approach and other techniques is presented in Section 6. Eventually, the effectiveness of the method, especially for small training set sizes, is discussed in Section 7.

## 2. Scale Space Texture Classification

A texture classification system typically consists of several stages like preprocessing, feature extraction, and classification, which are discussed in this and next sections.

### 2.1. Construction of Multiscale Textures

In the recent years, multiresolution techniques gained importance in texture analysis due to the intrinsic multiscale nature of textures. Among the multiscale techniques in the literature summarized in Section 1.1, the techniques based on scale space theory provide a mathematical framework for texture analysis which is biologically motivated by the models of the early stages of human vision [29]. We express the theory of scale space in the context of texture classification as follows:

We assume here that we only deal with gray scale textures. A texture can be considered as a function $L_i$ that maps the spatial information into intensity levels, i.e.,

$$L_i : \mathbb{Z}^2 \to \mathbb{Z}, \quad i = 1, ..., c \tag{1}$$

where $c$ is the number of textures (classes) to be classified. Spatial information and intensity are both assumed to be quantized[1].

According to scale space theory [29], to extract the multiscale differential texture structure $\partial^n L_i(\mathbf{x}; \sigma)/\partial \mathbf{x}^n$, $i = 1, ..., c$, $n \geq 0$, where $\sigma$ is the scale, we need to calculate the derivative of the observed image by a Gaussian aperture:

$$\frac{\partial^n}{\partial \mathbf{x}^n} L_i(\mathbf{x}; \sigma) = \frac{\partial^n}{\partial \mathbf{x}^n}(L_i * G_\sigma)(\mathbf{x}), \tag{2}$$

---

[1] In fact, the formulation provided for scale space can be applied to continuous images as well. However, the digital images are always quantized in both spatial and intensity spaces and hence we have assumed this as a realistic situation. Scale space addresses taking the derivative of discrete data (like digital images) by applying regularization on the image (by convolving image with a Gaussian kernel). This is usually addressed in the field of image processing by approximating differentiation by differencing which is avoided here.

where $*$ is the convolution operator and $G_\sigma$ is the 2D Gaussian kernel at scale $\sigma$

$$G_\sigma(\mathbf{x}) = \frac{1}{2\pi\sigma^2} e^{-\frac{\mathbf{x}^2}{2\sigma^2}}. \tag{3}$$

Since both convolution and differentiation are linear operators in (2), they can be commuted

$$\frac{\partial^n}{\partial \mathbf{x}^n} L_i(\mathbf{x};\sigma) = (L_i * \frac{\partial^n}{\partial \mathbf{x}^n} G_\sigma)(\mathbf{x}), \tag{4}$$

which means that to obtain the multiscale textures $\partial^n L_i(\mathbf{x};\sigma)/\partial \mathbf{x}^n$, $i = 1,...,c$, $n \geq 0$, one needs to convolve the original textures $L_i(\mathbf{x})$, $i = 1,...,c$, with the derivatives of the Gaussian kernel at multiple scales ($n = 0$ is for the convolution with the Gaussian kernel, i.e., for the zero$^\text{th}$ order derivative).

The order of the derivatives determines the type of structure to be emphasized. For example, the first order derivative extracts the edges, the second order emphasizes on the ridges and corners and so on. The order of the derivatives used to construct the multiscale textures might be application dependent. To avoid excessive computational load typically up to the second order derivatives are used. To address the issue of the number of orientations required in each derivative order, we use the steerability property of Gaussian derivatives [29, 37] for the efficient computation of directional derivatives. Based on this property, in the $n^\text{th}$ order derivative, $n + 1$ independent orientations are needed with which the derivatives in any orientation (in the same order) can be calculated. For example, for the first order derivative, i.e., $n = 1$, the derivative in arbitrary orientation $\theta$ can be calculated using $L_x$ and $L_y$:

$$L_\theta(x,y) = \cos(\theta) \times L_x(x,y) + \sin(\theta) \times L_y(x,y), \tag{5}$$

where $L_\theta$ is the derivative of $L(\mathbf{x})$ in orientation $\theta$.

As a general setup is considered in this research, the formulation provided in (4) is sensitive to orientation transformations. If rotation invariance is desired in an application, differential invariant descriptors, which are directly derived from combination of components of the local jet in scale space theory, can be adopted [38].

2.2. Construction of Multiscale Feature Spaces

To construct the multiscale feature space out of the multiscale textures, features are extracted from each scale to generate $n$-dimensional vectors $\mathbf{v}^{(i)} = [v_1,...,v_n] \in \mathbb{R}^n$, $i = 1,...,ns \times nd$, where $nd$ is the number of derivatives and $ns$ is the number of scales in each derivative. This generates $m = ns \times nd$ feature subsets at different scales/derivatives. Although this generation of feature subsets discussed in the context of construction of multiscale textures using scale space theory, it can be easily extended to other multiresolution techniques where $m$ feature subsets are generated at various resolutions. In other multiresolution techniques, $m$ also could be a multiplication of two other parameters, e.g., using Gabor kernels, the multiresolution textures are obtained by varying the center frequency in each band and the scale. Hence, $m$ will be the multiplication of the number of central frequencies and the number of scales. In this paper, however, we restrict ourselves to Gaussian derivatives for the generation of feature subsets. These feature subsets can be composed into a single feature vector $\mathbf{v} = [\mathbf{v}^{(1)}, \mathbf{v}^{(2)},...,\mathbf{v}^{(m)}]^\top$, which is called the *distinct pattern representation* (DPR) [39]. The dimensionality of the various feature subsets $\mathbf{x}(i)$, $i = 1, ..., m$, is not necessarily the same. As we will discuss later in Section 2.5, fewer features will be extracted from higher scales due to the coarser structures or less information available at these scales.

*2.3. Combined Classifiers versus Combined Feature Spaces*

After extraction of feature subsets $\mathbf{v}^{(i)} = [v_1, ..., v_n]^\top \in \mathbb{R}^n$, $i = 1, ..., m$, from all scales, the common trend in the literature is to concatenate these feature subsets to build up a single feature space $\mathbf{v} \in \mathbb{R}^{n \times m}$, called *combined feature space* (CFS) in this paper, where $n$ is the dimensionality of each feature subset and $m$ is the total number of feature subsets. The dimensionality of different feature subsets ($n$) might not be necessarily the same. However, for the time being and for the simplicity of the discussion, we assume that all feature subsets have the same dimensionality. The obtained fused feature space is then submitted to a single classifier $D : \mathbb{R}^{n \times m} \to \Omega$, where $\Omega = \{\omega_1, ..., \omega_c\}$ is the set of class labels for the textures.

The main disadvantage of this approach is that the fusion of feature subsets generates a high dimensional feature space that may cause suffering from the 'curse of dimensionality' [32]. This can be solved in three ways:

1. By reducing the dimensionality of each feature subset significantly. This can be done by a general procedure like principal component analysis (PCA), derived from the data, or based on the specific nature of the features. For example, in filter bank approaches like Gabor or wavelets, the outputs are not directly used. The features are produced by taking the local standard deviation, or the local energy, i.e., by a kind of energy estimate for each local window/patch extracted around each pixel [5, 19]. Some researchers calculate the moments of histogram on patches or regions of interest (ROIs) extracted from multiscale textures [10, 27, 28].
2. By using classifiers that behave well in high dimensional feature spaces like support vector machines (SVMs) [9, 24] a significant feature reduction in each scale is not needed.
3. By submitting the DPR to an ensemble of classifiers

$$\wp = \{D_1, ..., D_m\}, \quad \wp : \mathbb{R}^{n \times m} \to \Omega^m, \tag{6}$$

   where $D_i : \mathbb{R}^n \to \Omega$, $i = 1, ..., m$, is the base classifier trained on each feature subset $\mathbf{v}^{(i)} \in \mathbb{R}^n$, $i = 1, ..., m$. Then the decisions made by these BCs are fused. Thus, the problem of finding a classifier $D : \mathbb{R}^n \to \Omega$ is converted into finding an aggregation method $\Im : \Omega^m \to \Omega$ for combining the classifier outputs.

Our focus is on the third approach where we construct a two-stage combined classifier in the scale space texture classification context. We, thereby, emphasize more on the classifier structure than on the multiresolution technique.

*2.4. The Construction of DPR for the Ensemble of Classifiers*

In the above methods, the features are usually calculated on local windows/patches around pixels extracted from the constructed multiscale textures. To reduce the dimensionality of the feature space in the CFS approach, usually the energy of output filters, the moments of histograms, or other measurable parameters are calculated from these local patches and are considered as the feature subset $\mathbf{v}^{(i)}$, $i = 1, ..., m$ for a particular scale. These feature subsets are then concatenated to construct the fused feature space **v** to be submitted to a single classifier.

In the proposed approach using combined classifiers, the feature space obtained from each scale is submitted to a BC. Instead of fusion of features from different scales, the decisions made by the BCs are combined. Thus, a severe feature reduction is not required. Hence, in this approach, to construct the multiscale feature space (or the DPR), the pixels from the extracted patches in each scale can be used to construct the feature subsets in that scale. This means that each feature subset from each scale is constructed using the pixels in a patch extracted from this scale. As many patches as necessary are extracted to construct the training and test sets. The dimensionality of this feature subset ($n$) depends on the number of pixels in the patch, i.e., patch size.

To address the issue related to how to extract the patches from different scales we notice that the patch size is scale dependent and at higher scales (lower resolutions), the patch size should be increased. This is mainly because as we go to higher scales, more emphasis is given on coarser structures and hence they should be looked at through larger windows. On the other hand, the patches from different scales/derivatives of the same texture must be extracted from the same spatial locations to identify to what extent the additional information provided by other scales/derivatives can improve the performance of the classification system.

## 2.5. Subsampling and Feature Reduction

As explained above, at higher scales, the patch sizes should be increased to represent the coarser structures available at these scales. To overcome the resulting problem of large memory storage and high computational costs, subsampling can be performed at higher scales. In [40], it is shown that subsampling will not degrade the performance of the multiscale texture classification system while it can reduce the required memory space and computational cost.

To reduce the dimensionality of feature subsets further, feature extraction/selection techniques can be deployed, such as *principal component analysis* (PCA) or *independent component analysis* (ICA) [41]. While PCA is the most prevalent feature extraction method in the literature, there have recently been several works published to compare PCA and ICA as the feature extraction techniques in object recognition with some contradictory results (refer to [42] for a list). In Chapter 7 of [43], it is concluded that it is very difficult to find the independent components in a very high dimensional feature space and it usually leads to no significant improvement over PCA. In [42], it is shown how PCA and ICA are related as the feature extraction techniques and under which conditions their performance is completely equivalent. There are several suggestions to get better performance by ICA than by PCA where using feature selection (a subset of the original feature space to avoid searching in a very high dimensional feature space, which is consistent with the conclusion in [43]) is one main possibility [42]. Considering the simplicity of PCA in comparison to ICA and to avoid an extra feature selection step, which is not very obvious in our task, PCA is adopted in this research for feature extraction. Its application to scale space texture classification using combined classifiers is discussed in [40, 44]. PCA performs an adaptive feature extraction in multiscale texture classification in this sense that at higher scales the size of the patches need to be increased and hence more dimension reduction is required. PCA adapts itself according to scale as there are fewer details at higher scales and thus fewer components are needed to maintain the same fraction of variance of the original space.

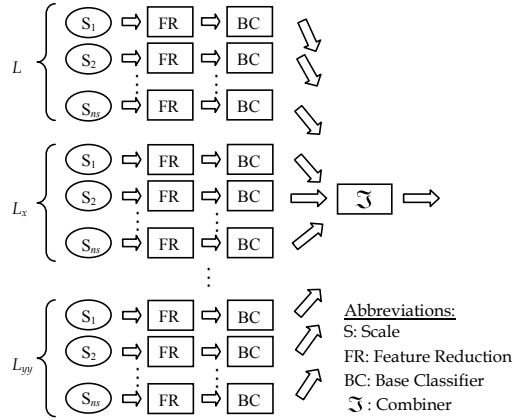

Figure 1: The structure of multiscale texture classification using one-stage combined classifier proposed in this paper.

By applying subsampling and PCA, the feature subsets $\mathbf{v}^{(i)}$, $i = 1, ..., m$ are mapped to an uncorrelated space. This builds up new feature subsets $\mathbf{u}^{(i)}$, $i = 1, ..., m$, which yields a new DPR $\mathbf{u} = [\mathbf{u}^{(1)}, \mathbf{u}^{(2)}, ..., \mathbf{u}^{(m)}]^\top$. The dimensionalities of new feature subsets might be different from each other and depend on the number of components needed by PCA to retain specific fraction of variance of the original space. The dimensionality of feature subsets at coarser scales is usually much lower than the finer scales [40].

Fig. 1 shows the overall structure of the scale space texture classification system using combined classifiers designed so far.

## 3. Combined Classifiers in the Context of Multiscale Texture Classification

As described in the previous section, instead of fusion of feature subsets obtained from different scales/derivatives, it seems to be natural to submit each feature subset to one classifier and combine their decisions.

The types of combined classifiers are: stacked, which is used in applications with the same feature space; parallel, with application in different feature spaces; and sequential, where the classifiers are applied sequentially one after another. As the features subsets generated from different scales/derivatives are different in our method, parallel combined classifiers seem to be the natural choice.

There are four levels of design related to the construction of the classifier ensembles [45], i.e., *data level*: different datasets to train the BCs; *classifier level*: the different BCs in the ensemble; *combination level*: the construction of combiners for the decisions made by the BCs; and *feature level*: different feature subsets might be applied to the BCs and the ensemble. In this paper, our focus is mainly on combination and feature levels.

For the second level of design, i.e., the classifier level, a variety of parametric classifiers, e.g., the linear discriminant classifier (LDC) and the quadratic discriminant classifier (QDC), as well as nonparametric classifiers, e.g., the $k$ nearest neighbor ($k$-NN) and Parzen classifiers are examined among which QDC performed the best and hence selected as the BC.

In this paper, our focus is mainly on combination and feature levels. In feature level, although the feature subsets are always generated using multiscale differential operators to construct multiscale textures, different combinations of the feature subsets are mainly investigated. This is different from the combination level, where our focus is on the type of combiner for the outputs of the classifier ensemble. These two different levels are discussed in the next subsections in more details.

*3.1. Combination Level - The Type of Combiner*

Each base classifier $D_i$, $i = 1, ..., m$, in the ensemble $\wp$ generates a $c$-dimensional vector $[d_{i,1}, ..., d_{i,c}]^\top$. Two types of outputs are produced by the classifier ensemble[2]:

1. Class labels. In this case, we can assume that the output of each base classifier $D_i$, $i = 1, ..., m$, in the ensemble $\wp$ generates a $c$-dimensional binary vector $[d_{i,1}, ..., d_{i,c}]^\top \in \{0, 1\}^c$, where $d_{i,j} = 1$ if $D_i$ labels $\mathbf{u}$ in class $\omega_j$, and 0 otherwise.
2. Continuous values, i.e., each base classifier $D_i$, $i = 1, ..., m$, in the ensemble $\wp$ generates a $c$-dimensional vector $[d_{i,1}, ..., d_{i,c}]^\top \in [0, 1]^c$, where each $d_{i,j}$ represents the support for the hypothesis that the DPR $\mathbf{u}$ submitted for classification to the classifier ensemble belongs to the $\omega_j$, $j = 1, ..., c$. This can be, for example, the posterior probability $p(\omega_j|\mathbf{u})$. In this way, the whole classifier ensemble generates a matrix of $d_{i,j}$, $i = 1, ..., m$, and $j = 1, ..., c$, which is called a decision profile $\boldsymbol{DP}(\mathbf{u})$ [45]:

$$\boldsymbol{DP}(\mathbf{u}) = \begin{bmatrix} d_{1,1} & \cdots & d_{1,j} & \cdots & d_{1,c} \\ \vdots & \ddots & \vdots & \ddots & \vdots \\ d_{i,1} & \cdots & d_{i,j} & \cdots & d_{i,c} \\ \vdots & \ddots & \vdots & \ddots & \vdots \\ d_{m,1} & \cdots & d_{m,j} & \cdots & d_{m,c} \end{bmatrix} \quad (7)$$

where the elements $d_{i,j}$ of matrix $\boldsymbol{DP}(\mathbf{u})$ are also dependent on $\mathbf{u}$. However, this dependency is not shown in (7) and in the text for the simplicity of notation. In (7), each row is the output of classifier $D_i(\mathbf{u})$ and each column is the support from BCs, $D_i$, $i = 1, ..., m$, for class $\omega_j$.

The main goal at the combination level is to design an aggregation function $\Im$ to combine the decisions at the output of BCs into a single decision. The design of $\Im$ depends on the type of the outputs of BCs.

*3.1.1. Combiners Based on the Class Labels*

Here, the aggregation function $\Im$ selects the overall class label of the ensemble $\delta_\wp(\mathbf{u})$ based on the class labels generated at the output of each BC in the ensemble. The most common combiner based on class labels is majority voting. This combiner considers the

---
[2]PRTools is used for the Matlab implementation of the algorithms [46].

output of the ensemble as the class label appeared at the output of the BCs more than others:

$$\delta_\wp = \{\delta_k | k = \arg\max_{j=1}^{c} \sum_{i=1}^{m} d_{i,j}\}. \tag{8}$$

*3.1.2. Combiners Based on the Continuous Outputs*

The first step to use the continuous valued outputs of the BCs (e.g. confidences) is to make sure that they are normalized so all the outputs have equal contribution towards the overall performance of the system. Consequently, the confidences are summed to one at the outputs of the base classifiers and can be considered as posterior probabilities. In other words, the sum of the elements on each row of $\boldsymbol{DP}(\mathbf{u})$ matrix is one after normalization. This, however, does not alter the class membership of the data samples submitted for classification in the maximum a posteriori (MAP) probability sense. Here, we assume that $d_{i,j}$ values of $\boldsymbol{DP}(\mathbf{u})$ matrix given in (7) are normalized.

Combiners based on continuous valued outputs of the BCs are divided into two groups: nontrainable and trainable combiners.

In *nontrainable combiners*, the aggregation function $\Im$ calculates the support for the hypothesis that the submitted DPR $\mathbf{u}$ belongs to class $\omega_j$, using only the $j^{\text{th}}$ column of $\boldsymbol{DP}(\mathbf{u})$ by

$$\mu_j(\mathbf{u}) = \Im(d_{1,j}, ..., d_{m,j}). \tag{9}$$

There is no additional parameter to train. As soon as the outputs of the BCs are available, the overall output of the combined classifier can be calculated.

The most common nontrainable combiners are:

$$
\begin{array}{lll}
\text{min combiner} & : & \mu_j(\mathbf{u}) = \min_{i=1}^{m} d_{i,j} \\
\text{prod combiner} & : & \mu_j(\mathbf{u}) = \prod_{i=1}^{m} d_{i,j} \\
\text{median combiner} & : & \mu_j(\mathbf{u}) = \underset{i=1}{\overset{m}{\text{median}}}\, d_{i,j} \\
\text{mean combiner} & : & \mu_j(\mathbf{u}) = \frac{1}{m}\sum_{i=1}^{m} d_{i,j} \\
\text{max combiner} & : & \mu_j(\mathbf{u}) = \max_{i=1}^{m} d_{i,j}
\end{array}
\tag{10}
$$

Among these combiners, the min combiner selects the class to which all classifiers object the least. It is thereby careful and not sensitive for possible overtrained classifiers. The max combiner selects the class chosen by the most confident classifier. This is good if all classifiers are well trained and none of them is overtrained. The other combiners are between these two extremes. The mean and product combiners are the most common ones in the literature and their properties have been more investigated. While it is still not clear which combiner is the best, the results from other researches show that mean and product combiner may perform well in many applications [45]. However, according to experiments done by Kittler et al. [39], the mean combiner is more resilient to noise than the product combiner.

Among *trainable combiners*, decision templates (DTs) are the simplest one. The DT can be calculated by averaging the $\boldsymbol{DP}(\mathbf{u})$ for the whole DPR. Then by calculating the decision profile for a test object and finding the nearest distance from the DT to this DP, the class label for the test object can be decided.

*3.2. Feature Level - Combining of Feature Subsets*

In the construction of multiscale textures, different scales and different derivatives are used and hence, the feature subsets in DPR come from different scales of different derivatives. Each feature subset $\mathbf{u}^{(i)}$ is submitted to one BC. Combining the BCs can be done in two ways: by combining all of them in one stage, as already explained and illustrated in Fig. 1; and by combining in two stages.

In the latter case, there are several ways of combining. However, two straightforward methods of combinations are:

1. Different scales of the same derivative in the first stage and different derivatives in the second.
2. Different derivatives at the same scale in the first stage and different scales in the second.

It is also possible to combine different scales of different derivatives in various ways. However, the above methods of combining group the feature subsets in a natural way, i.e., different derivatives of the same scale or different scales of the same derivative. Since the paper is about multiscale texture classification, in presenting the results, we focus more on combining different derivatives of the same scale to better realize the contribution of different scales towards the overall performance of the system.

Moreover, it is also possible that a technique between combined classifiers and CFS is used: Concatenation of feature subsets in the first stage, then submitting them to the BCs and combining their decisions in the second stage. There are again various ways of combination. However, two more natural possibilities are:

1. Fuse feature subsets from different scales of the same derivative in the first stage and combine the decisions made by the BCs in the second stage, or
2. Fuse feature subsets from different derivatives at the same scale in the first stage and combine the decisions made by the BCs in the second stage.

The architectures of two-stage scale space texture classification using combined classifiers for each of the above combinations are shown in Fig. 2.

*3.2.1. A Two-Stage Combiner: Different Scales in the First Stage and Derivatives in the Second*

First we show the formulation for the two-stage combined classifier shown in Fig. 2a, i.e., combining the outputs of BCs for different scales of the same derivative in the first stage and then different derivatives in the second stage. The starting point is the matrix $\boldsymbol{DP}(\mathbf{u})$ in (7). However, we need to rearrange it as shown in Fig. 3a, where $ns$ is the number of scales and $nd$ is the number of derivatives. The decision profile $\boldsymbol{DP}(\mathbf{u})$ in Fig. 3a is subdivided into some submatrices by the dashed lines. Each submatrix is generated from the outputs of the same derivative at different scales in the first stage of the classifier ensemble (see Fig. 2a).

Two aggregation functions are needed here. One aggregation function $\Im$ to combine the elements of one column in each submatrix and the second one $\varphi$ to combine all submatrices, i.e.,

$$\mu_j(\mathbf{u}) = \varphi[\Im(\mathbf{d}_0^j), \Im(\mathbf{d}_1^j), ..., \Im(\mathbf{d}_{nd-1}^j)], \tag{11}$$

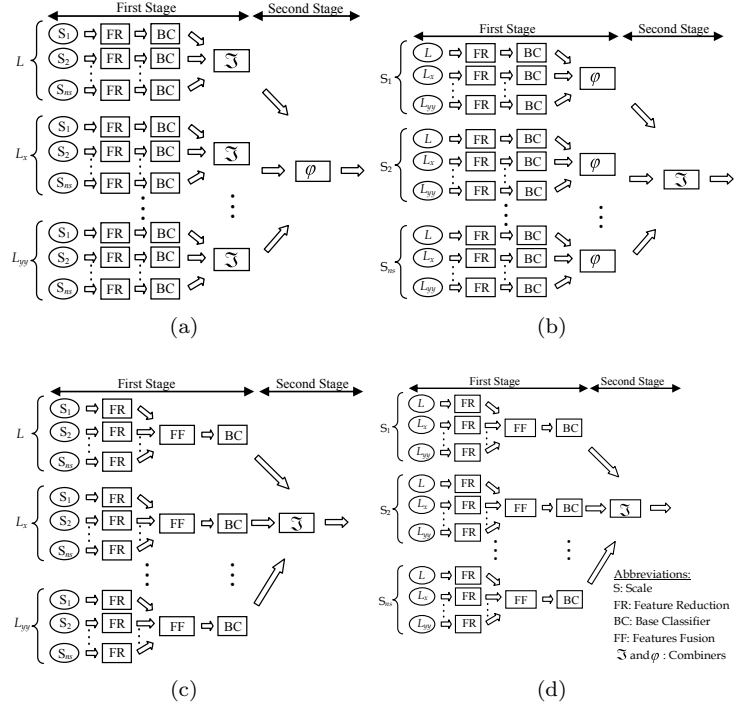

Figure 2: The structure of multiscale texture classification using two-stage combined classifier proposed in this paper with (a) combining different scales of the same derivative in the first stage and different derivatives in the second stage, (b) combining different derivatives at the same scale in the first stage and different scales in the second stage, (c) fusion of feature subsets obtained from different scales of the same derivative in the first stage and combining BCs in the second stage, and (d) fusion of feature subsets obtained from different derivatives at the same scale in the first stage and combining BCs in the second stage.

where $\Im : [0,1]^{ns} \to \mathbb{R}$ is the first aggregation function and the vector $\mathbf{d}_k^j$ is defined as

$$\mathbf{d}_j^k = (d_{1+k\times ns,j}, d_{2+k\times ns,j}, ..., d_{ns+k\times ns,j}), \quad k = 0, ..., nd-1. \tag{12}$$

If both $\Im$ and $\varphi$ are of the same type then for all of the nontrainable combiners in (10) except for the median combiner, $\mu_j(\mathbf{u})$ from (11) is the same as what is obtained from one stage combiner using the corresponding aggregation function as given in (9). For example, if both combiners are mean combiners then

$$\mu_j(\mathbf{u}) = \frac{1}{ns \times nd} \sum_{k=0}^{nd-1} \sum_{i=1}^{ns} d_{i+k\times ns,j}, \tag{13}$$

and since $m = ns \times nd$, this is equivalent to $\mu_j(\mathbf{u}) = (\sum_{i=1}^m d_{i,j})/m$, which is the mean combiner in one stage combined classifier given in (10). However, $\Im$ and $\varphi$ combiners are not necessarily the same. For example, the mean of product, i.e., $\Im$ as mean combiner and $\varphi$ as product combiner, can be formulated as follows:

$$\mu_j(\mathbf{u}) = \prod_{k=0}^{nd-1} \left(\frac{1}{ns}\sum_{i=1}^{ns} d_{i+k\times ns,j}\right). \tag{14}$$

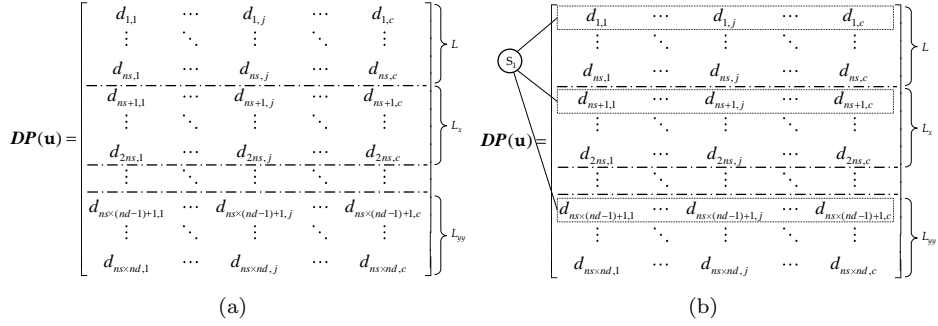

Figure 3: The decision profile given in (7) rearranged to group (a) different scales of the same derivative and (b) different derivatives of the same scale.

Other combinations can be formulated similarly.

*3.2.2. A Two-Stage Combiner: Different Derivatives in the first Stage and Scales in the Second*

The formulation of the combiner in this case (see Fig. 2b) is similar to what is shown before with some modifications. First, grouping has to be done in different way for decision profile matrix $\boldsymbol{DP}(\mathbf{u})$, as shown in Fig. 3b.

Combining the derivatives at the same scale first and then combine all the results obtained is similar to applying the aggregation function $\varphi$ used in (11) first and then apply $\Im$ defined in the same equation

$$\mu_j(\mathbf{u}) = \Im[\varphi(\mathbf{d}_1^j), \varphi(\mathbf{d}_2^j), ..., \varphi(\mathbf{d}_{ns}^j)], \tag{15}$$

where $\varphi : [0,1]^{nd} \to \mathbb{R}$ is the first aggregation function and the vector $\mathbf{d}_i^j$ is defined as

$$\mathbf{d}_i^j = (d_{i,j}, d_{i+ns,j}, ..., d_{i+ns\times(nd-1),j}), \quad i = 0, ..., ns. \tag{16}$$

Care must be taken that even if both aggregation functions $\Im$ and $\varphi$ are linear then the result of (15) will not be necessarily the same as what is obtained from (11) at the output of classifier ensemble. This is because, in general, the equation $\Im \circ \varphi = \varphi \circ \Im$, where $\circ$ is the composition operator on functions, does not hold. For example, if the same as what was formulated in (14), $\Im$ is mean combiner and $\varphi$ is product combiner then the output of combined classifier for Fig. 2b can be calculated using (15) as follows:

$$\mu_j(\mathbf{u}) = \frac{1}{nd}\sum_{k=0}^{nd-1}\left[\prod_{i=1}^{ns} d_{i+k\times ns,j}\right], \tag{17}$$

which is not equivalent to $\mu_j(\mathbf{u})$ obtained from (14).

*3.2.3. Fusion of Features in the First Stage and Combining Base Classifiers in the Second*

Concatenation of features in the first stage can be performed in two ways: fusion of features obtained from different scales of the same derivative or fusion of features obtained from the same derivative at different scales. After this fusion, obtained feature subsets can be applied to some BCs and a combiner can calculate the overall output of the system. Here, we only discuss the concatenation of features from different scales in the first stage and the discussion on the concatenation of features from different derivatives is similar.

The feature subsets obtained from fusion of features at different scales of the same derivative can construct a new DPR $\mathbf{z} = [\mathbf{w}^{(1)}, ..., \mathbf{w}^{(nd)}]$, where $\mathbf{z}$ is the new DPR constructed and $\mathbf{w}^{(i)}$, $i = 1, ..., nd$, are the fused feature subsets obtained from fusion of features at different scales of the same derivative. Each of these fused feature subsets $\mathbf{w}^{(i)}$, $i = 1, ..., nd$, is submitted to a BC and the decision profile $\boldsymbol{DP}(\mathbf{z})$ is as follows:

$$\boldsymbol{DP}(\mathbf{z}) = \begin{bmatrix} d_{1,1} & \cdots & d_{1,c} \\ \vdots & \ddots & \vdots \\ d_{nd,1} & \cdots & d_{nd,c} \end{bmatrix}. \tag{18}$$

The combiner is applied to $d_{i,j}(\mathbf{z})$ of each column

$$\mu_j(\mathbf{z}) = \Im(d_{1,j}, ..., d_{nd,j}). \tag{19}$$

For example, the mean combiner is

$$\mu_j(\mathbf{z}) = \frac{1}{nd} \sum_{i=1}^{nd} d_{i,j}. \tag{20}$$

As before, normalization has to be applied before the combiner to ensure $d_{i,j} \in [0, 1]$.

## 4. Experiments

To evaluate the performance of the proposed system in the classification of small image patches, a variety of tests are performed on a supervised classification of some test texture images from standard databases.

*4.1. Dataset*

Brodatz album [47] is used to evaluate the performance of the proposed texture classification system. The textures used from Brodatz album are shown in Fig. 4. All textures in these texture combinations are homogeneous and have a size of $640 \times 640$ pixels with 8 bit/pixel intensity resolution. They are historically selected based on the experiments made in [24] for 4-class and in [19] for 5-, 10-, and 16-class problems to make the initial comparison easier. However, the comparison with the results in these two papers will not be provided in Section 6 despite of the superiority of our results. This is mainly because the test conditions and experimental set up are different and comparing the results directly may not be fair.

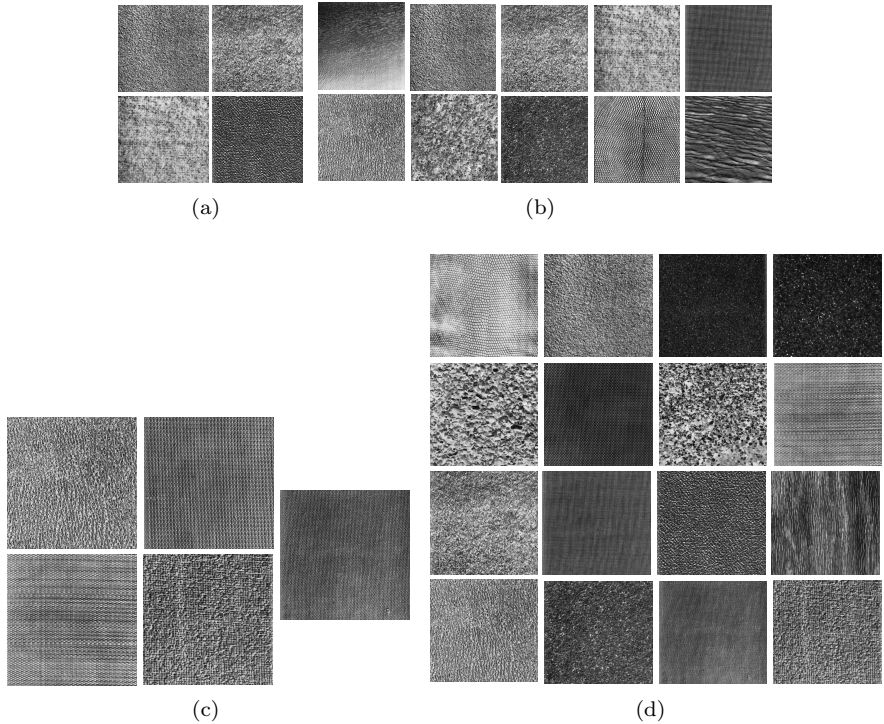

Figure 4: The texture images from Brodatz album used in the experiments with (a) 4-class including D4, D9, D19, and D57, (b) 5-class including D24, D53, D55, D77, and D84, (c) 10 class including D4, D9, D19, D21, D24, D28, D29, D36, D37, and D38, and (d) 16-class consisting of D3, D4, D5, D6, D9, D21, D24, D29, D32, D33, D54, D55, D57, D68, D77, and D84 textures.

### 4.2. Data Preparation and Patch Extraction

As there is only one texture in each class in Brodatz album, the training and test sets are generated from the same image in each scale/derivative. Hence, they may spatially overlap in the image domain. To ensure that they are fully separated, the texture images are split spatially into an upper and lower half. Training and test sets are generated from these halves respectively.

Further preparation for the experiments involves preprocessing, the computation of multiscale textures, and the extraction of patches. These three steps can be done in different orders. To prevent the influence of neighboring pixels on the computation of multiscale textures of each patch and to make this data preparation more realistic[3], we first extract the patches and then preprocess them and construct the textures. The patches have size 32 × 32 pixels. In Brodatz, to generate sufficient patches for training and testing, the patches have some overlap. These patches are extracted from each half separately from top left to bottom right and in total 1769 patches are extracted from each half.

---

[3]In real world data, we usually have a region of interest (ROI) on which the multiscale textures are calculated. Each ROI is considered as a sample in train or test set and they do not influence each other during the convolution process required for multiscale computation.

*4.3. Preprocessing*

Preprocessing is performed on the patches separately. We made the patches indiscriminable to the mean and variance by DC cancellation and variance normalization before the construction of multiscale textures. This guarantees that the textures are indiscriminable to the average intensity level and contrast.

*4.4. Computation of Multiscale Textures*

The N-jet of scaled derivatives up to the second order is used to compute multiscale textures on each patch. This yields $L$, $L_x$, $L_y$, $L_{xx}$, $L_{xy}$, and $L_{yy}$ at multiple scales. The convolution of patches of 32 × 32 pixels with Gaussian derivatives are needed as given in (4). Some preliminary experiments showed that using cyclic or reflective techniques on the boundary of the patches for the computation of this convolution did not have any effect on the results. Hence, the reflective technique is used at the boundaries.

The derivatives are to be calculated at some scales as given in (4). There are two important issues here, the number of scales, and the scales at which the derivatives are to be calculated, the working scales. The working scale is texture dependent, i.e., for finer textures lower scales and for coarser textures higher scales can be used. The number of scales is application dependent and affects the computational load of the algorithm. Practically, only three scales are included.

The working scales can be found using algorithms that specify the keypoints in the textures like scale invariant feature transform (SIFT) [48] and top-points [49]. This means that above certain scales, textures will not be discriminative and will not contribute to the performance of the overall system. Based on this observation, the Gaussian derivatives are calculated at three scales, i.e., at scales $S_1$, $S_2$, and $S_3$ with the variance ($\sigma^2$) of 1, 4, and 7 respectively. These scales are chosen based on some preliminary experiments that showed they provide sufficiently different but still informative results. Hence, out of each patch, 6 (derivatives) × 3 (scales) texture patches are obtained.

*4.5. Construction of Training and Test Sets*

As mentioned in Sections 4.2 to 4.4, in Brodatz album, the patches for training and testing are extracted from different image halves, preprocessed and then multiscale textures are calculated on each patch. As discussed in Section 2.4 and 2.5, in the proposed approach, the pixels in a patch are used to construct the feature subsets at every scale and derivative. There are 1769 patches of 32 × 32 at each scale/derivative for training and the same number of patches for testing. These patches are randomly ordered and then as many as needed are used for training and testing.

Different patch sizes at different scales are used in the experiments. Increasing patch size with scale can improve the performance of the system especially for larger training set sizes. However, due to peaking phenomenon[4], it degrades the performance in smaller training set sizes and increases the computational load. So, as a compromise, in all experiments the patch sizes of $18 \times 18$, $24 \times 24$, and $30 \times 30$ are used at scales $S_1$, $S_2$, and $S_3$ respectively. These patches are taken from central part of multiscale patches. Test set size is fixed at 900. Training set size is increased from 10 to 1500 to construct the learning curves and to investigate the performance of the classification system at various training set sizes.

---

[4]The peaking phenomenon is related to this observation: while increasing the number of features may first decrease the classification error, it will start degrading the performance of a classifier at certain point if the training set size is not increased [32].

### 4.6. Feature Extraction

PCA is used for feature extraction in the proposed approach and it is calculated over all classes in each scale/derivative separately. The number of components retained at the output of PCA, which in fact determines the dimensionality of each feature subset $\mathbf{u}^{(i)}$, $i = 1, ..., m$ in transformed (uncorrelated) DPR $\mathbf{u} = [\mathbf{u}^{(1)}, \mathbf{u}^{(2)}, ..., \mathbf{u}^{(m)}]^\top$, is selected to preserve 95% of the variance in original feature subsets $\mathbf{v}^{(i)}$, $i = 1, ..., m$ of DPR $\mathbf{v} = [\mathbf{v}^{(1)}, \mathbf{v}^{(2)}, ..., \mathbf{v}^{(m)}]^\top$.

### 4.7. Classifier

One- or two-stage combined classifier is used. Among the BCs tested, quadratic discriminant classifier (QDC) performs the best and hence used in most of the experiments. One BC is used for every scale in one derivative. Care must be taken that even after applying PCA, the number of features retained at scale one is rather high (typically more than 150 components [40]). This degrades the performance of the BCs at this scale and thereby the overall performance of the classification system at small training set sizes. Hence, two regularization parameters are used for the estimation of covariance matrix in QDC [46]

$$\hat{\Sigma} = (1 - \eta - \lambda)\Sigma + \eta(\Sigma \odot \mathbf{I}_n) + \lambda\frac{\mathrm{tr}(\Sigma)}{n}\mathbf{I}_n, \tag{21}$$

where $\Sigma$ is the class covariance matrix, $\eta$ and $\lambda$ are regularization parameters, $\mathbf{I}_n$ is the identity matrix, $\odot$ denotes elementwise product (the ".*" notation of Matlab), and $n$ is the dimension of $\Sigma$. A fixed regularization is performed at scale one with both parameters $\eta$ and $\lambda$ set to 0.01.

The output of each BC in the first stage is normalized before applying the combiner at this stage to ensure that all outputs $(d_{i,j})$ from different BCs have the same amount of significance on the overall output of the combiner.

### 4.8. Evaluation

One of the main shortcomings of the papers in multiresolution texture classification is reporting the performance of the algorithm for one single training set size. This causes that the behavior of the algorithm for different training set sizes remains unrevealed.

In this paper, to overcome this problem and to investigate the performance of the classifier, the learning curves are drawn by calculating the error for different training set sizes up to 1500. The experiments are repeated 5 times, each time using different set of random patches extracted from the multiscale textures, and the errors calculated are averaged.

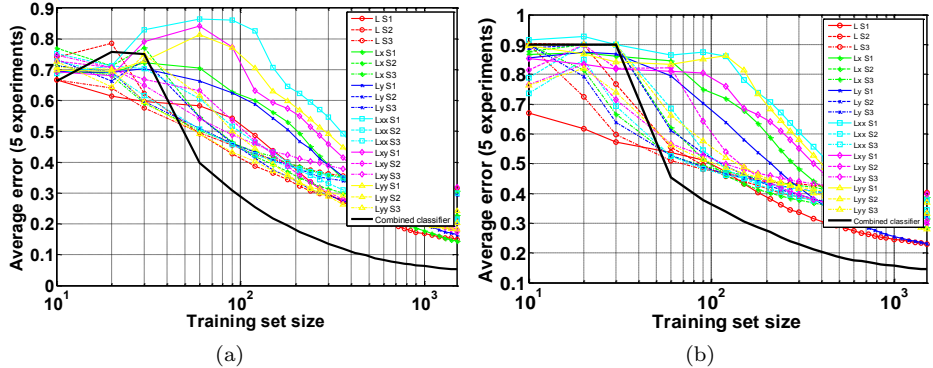

Figure 5: The performance of one-stage combined classifier on (a) 4-class of Brodatz (Fig. 4a) and (b) 10-class of Brodatz (Fig. 4b)

## 5. Results

A variety of tests are performed to evaluate the performance of the proposed multiscale texture classification system using combined classifiers under different conditions such as different parameters and input textures. To keep the paper short, only the most important results, which are essential in understanding this paper are included and many more can be found in the supplementary material provided.

Fig. 5 shows the learning curves for classification using one-stage combined classifier (refer to Fig. 1 for the structure of the classification system). The mean combiner is used and regularization is performed on QDC at scale $S_1$ for all derivatives. It clearly shows that the combined classifier (thick curve) performs much better than every single scale/derivative. This is especially significant in small training set sizes. For example, for a training set size of 100, improvements of more than 15% in the 4-class Brodatz problem and about 10% in the 10-class problem can be noticed over the best BC. For very small training set sizes (below 60), the error from each scale/derivative is still too high for the multiple classifier system to be useful. The poor performance of BCs at these very small training set sizes is due to peaking phenomenon at scales $S_2$ and $S_3$ as the number of components after applying PCA is comparable to the number of patches used for training and no regularization is used at these scales. If required, we can resolve this problem by any of these approaches: 1) using voting-mean combiner instead of mean-mean combiner as further discussed in Subsection 5.4, (2) using fewer components of PCA to avoid overtraining at these data samples, or (3) regularization at scale 2 to overcome the rather high dimensional feature space (comparing to the number of data samples available for training) at this scale after using PCA.

### 5.1. Classifier Level Design

Fig. 6 shows the effect of the type of the BCs on the performance of the two-stage combined classifier. Both parametric and nonparametric density estimation classifiers are tested. The results shown in Fig. 6 are for 4-class problem (Fig. 4a) with QDC, $k$-NN, and Parzen classifiers as BCs. Regularization is performed at scale $S_1$ for QDC.

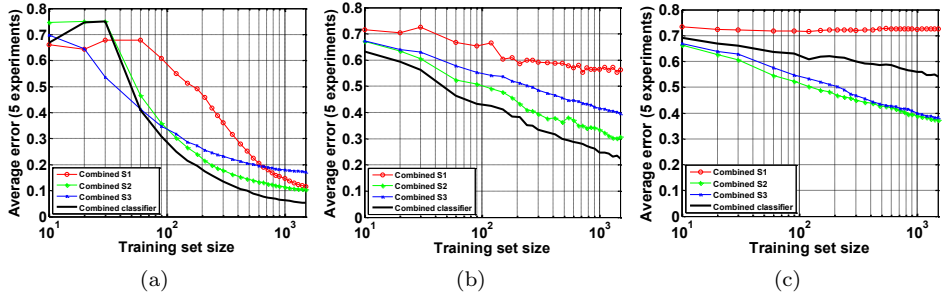

Figure 6: The effect of the type of BCs on the performance of scale space texture classification using two-stage combined classifier for 4-class problem of Brodatz album (refer to Fig. 4a). The BCs tested are: (a) QDC, (b) $k$-NN, and (c) Parzen. Thick curve is the overall combined classifier and thinner marked curves are the combined classifiers for different derivatives at the same scale (the outputs of first stage for the structure proposed in Fig. 2b).

The optimum value $k^*$ parameter in $k$-NN and $h^*$ (smoothing) parameter in Parzen is obtained for each scale/derivative using leave-one-out technique at each learning size. The structure of the system used is as what is shown in Fig. 2b, i.e., the system combines the different derivatives in the first stage and then combines the different scales in the second stage. Not shown are the results for LDC. As could be expected, its performance is almost the same as *prior* as we have removed the DC from the patches.

QDC performs the best, which may be related to the use of PCA for feature extraction. PCA is a linear dimension reduction technique. The projection of high dimensional data to lower dimensional subspaces normalizes the distributions as follows from the central limit theorem. Since the performance of QDC is the best, we use it as the BC in all experiments. Moreover, optimizing and using non-parametric procedures like the Parzen classifier is computationally expensive.

The results from other texture combinations are consistent with the results shown in Fig. 6 as shown in supplementary material.

## 5.2. The Effect of the Patch Size

Fig. 7a shows the performance of two-stage combined classifiers using 4-class problem of Brodatz album (Fig. 4a) in case that the patch size is the same at all scales (18 × 18). Comparing this to Fig. 6a reveals the importance of increasing patch sizes at higher scales.

The main reason as described before is the presence of coarser structures at higher scales. Hence, we need to look at these structures through larger windows. It can be seen from comparing Fig. 7a with Fig. 6a that the performance of the classifiers is degraded at higher scales, i.e., $S_2$ and $S_3$.

One may think that an alternative solution might be using the same patch size at all scales with the size of the patch according to the requirement from higher scales, i.e., 30 × 30. Fig. 7b shows the result of classification using this patch size at all scales for the same 4-class problem as in Fig. 7a and Fig. 6a. Increasing patch size at scales $S_1$ and $S_2$ to make it the same as scale $S_3$ causes that the dimensionality of feature subsets at these scales goes even higher and this degrades the overall performance of the system for small training set sizes as can be seen from comparing Fig. 6a and Fig. 7b. This also increases the computational load and memory storage. Although, this may improve the performance at larger training set sizes as a compromise among all these factors, i.e., the performance at both small and large training set sizes and the complexity, using smaller patch sizes at scale $S_1$ and larger ones at higher scales is concluded.

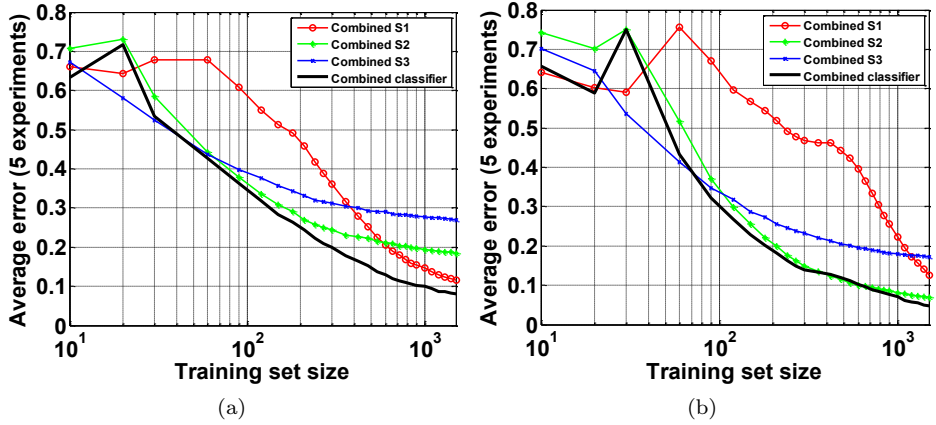

Figure 7: The effect of patch size on the performance of two-stage combined classifier applied to 4-texture problem of Brodatz for (a) patch size of $18 \times 18$ at all scales and (b) patch size of $30 \times 30$ at all scales.

### 5.3. Feature Level Design

Fig. 5a, Fig. 6a, and Fig. 8 show the learning curves for proposed structures of combined classifiers in scale space texture classification for 4-class problem of Brodatz album shown in Fig. 4a. Fig. 5a is one stage combined classifier using the mean combiner. Fig. 6a and Fig. 8 are the results for two-stage combined classifiers. In Fig. 6a, the different derivatives of the same scale are combined first and then different scales are combined in the second stage. In Fig. 8a, the different scales of the same derivative are combined and then different derivatives in the second stage. Mean combiners are used in both stages. As discussed in Section 3.2 and proved in (13), the overall performance in Fig. 5a, 6a and 8a are the same as the two-stage combined classifier performs the same as one-stage in this case. However, the intermediate results are different.

Fig. 8b and Fig. 8c show the learning curves for the combined classifier with the structures shown in Fig. 2c and Fig. 2d, respectively. In these two figures, feature subsets from different scales/derivatives are fused in the first stage and then presented to the BCs. In the second stage, the mean combiner combines the decisions made by these BCs. As can be seen, the performance is especially degraded for small to medium training set sizes. This is mainly because fusion of feature subsets in the first stage enlarges the dimensionality of the feature subsets applied to the BCs. This, in turn, makes the training of the system more difficult as more data samples are needed for the training of the BCs. Asymptotically, the system may perform the same as what is obtained from Fig. 6a or Fig. 8a. The conclusion is that the earlier the feature subsets are submitted to the BCs the better for small training sets.

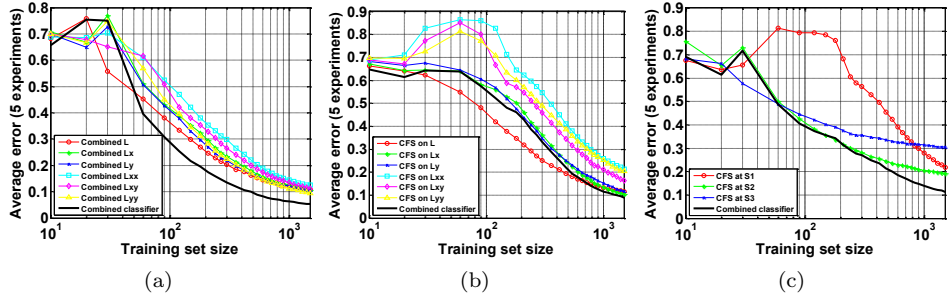

Figure 8: The effect of feature level design on 4-class problem of Brodatz: (a) two-stage combined classifier with combining scales in the first stage and derivatives in second stage with mean combiner at both stages (structure of Fig. 2a), (b) fuse feature subsets for different scales of the same derivative in first level and use combined classifiers with mean combiner in second level (structure of Fig. 2c), and (c) fuse feature subsets for different derivatives of the same scale in first level and use combined classifiers with mean combiner in second level (structure of Fig. 2d).

To address the issue of which derivatives are the most influential ones, we excluded one of the derivatives out each time and the results were degraded for less than 2%. Important remark here is that excluding the derivatives out degrades the results more severely at small training set sizes. Thus, with sufficient number of data samples, including the information from other derivatives become less important. We believe that the contribution of different derivatives also depends on the structure of the images. One can imagine that, for example, in a texture containing horizontal stripes, $L_y$ must have a large contribution whereas for a texture with vertical stripes, $L_x$ must be important. In general, since we do not know which structures are prevalent in the textures *a priori*, we use N-jet of derivatives up to specific order (in our paper 2) such that we include all possible discriminative structures. As for the scales, we started from 6 scales and reduced the number of scales to see how the results are affected. The results are not very much degraded when we reduce the number of scales to three. But, if we exclude scale 3 as well, then the results are significantly degraded. This is the reason that we used three scales in our experiments.

Here we just described the effect of method of combination of different feature subsets on the performance of the system. The results for using other types of combiners are shown in next subsection.

*5.4. Combination Level Design*

The results presented in this paper are all based on using the mean combiner for both stages. Hence, the overall results in Fig. 5a, 6a, and 8a are identical. We also experimented with other fixed and trainable combining rules like the product, maximum, minimum, median, voting, and decision templates. Their relative performances are dependent on the size of the training set as some are sensitive to overtrained classifiers. We found that, for instance, in very small training sets (less than 60) voting and the mean rule may perform better for the two-stage combiners. A larger study can be found in the supplementary material with many more experiments and details.

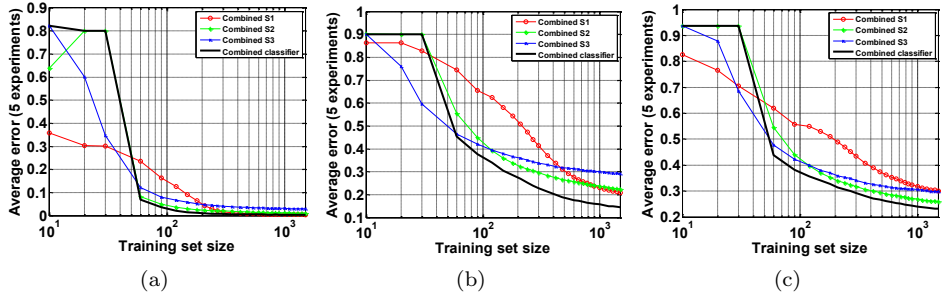

Figure 9: The results of two stage combined classifier with mean combiner in both stages (structure of Fig. 2b) for (a) 5-class, (b) 10-class, and (c) 16-class problems of Brodatz album shown in Fig. 4.

Since the mean-mean combiner performs the best consistently for all texture combinations used in this paper, the results for this combiner combination are shown in Fig. 9 for other textures from Brodatz. It shows that using the mean combiner at both stages also consistently improves the overall performance over each combined scale.

## 6. Comparison With Other Techniques

The performance of the scale space texture classification system using combined classifiers is compared against other techniques as follows. First we compare the proposed technique with the most similar ones and then try to compare it with more different techniques.

An important remark here is that the focus of this paper is on the classification of small patches not image classification as is done in the algorithms such as texton-based approach [50]. All our results are also presented in terms of patch classification (not image classification) error. Here, the goal is not image classification but it is small patch classification. In texton-based approach, many patches are extracted from each image (usually large) to build the histogram of textons as the model to represent this image. If the image itself is very small, the histogram of textons would have been too sparse to be reliable model for the image (see, for example, p. 977 of [12]).

### 6.1. Combined Classifier versus CFS

To demonstrate the superiority of combined classifier to CFS, a comparison between learning curves in Fig. 6a and Fig. 8c is very useful. The learning curves in Fig. 6a are obtained by submitting the feature subset at each single scale to a BC. Next, the decisions by these base classifiers are combined in two stages. Whereas, in Fig. 8c, the feature subsets from different derivatives of the same scale are first concatenated to construct a CFS and then each of these fused feature subsets is submitted to a BC. Eventually these BCs are combined. As can be seen for all scales, the learning curves in Fig. 6a demonstrate a better performance than those in Fig. 8c especially for small training set sizes. This is particularly shown for the zero$^{th}$ order derivative at multiple scales in Fig. 10.

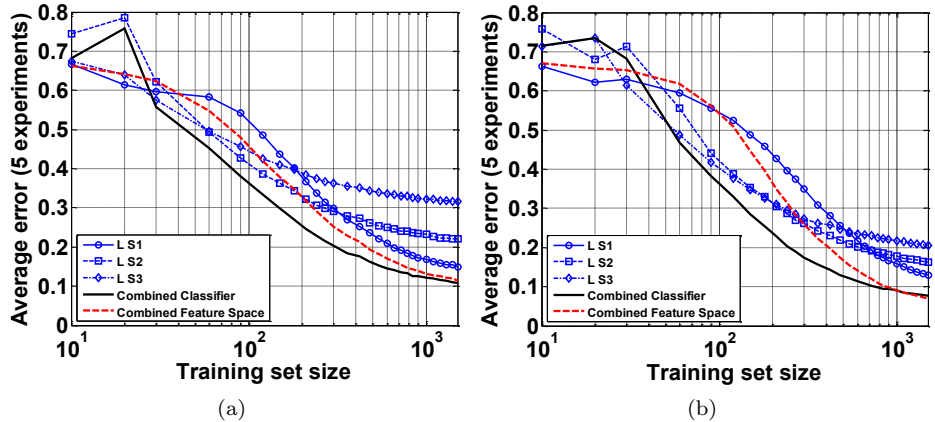

Figure 10: Combined classifier versus CFS for the zeroth order derivative on 4-class problem of Brodatz with regularization of BC at scale S1 for combined classifier approach and with (a) 95% and (b) 99% of retained variance for PCA in both approaches.

Fig. 10a compares combined classifiers and CFS for the case that PCA is used in such a way that 95% of the original variance is retained in the transformed space while Fig. 10b shows the same comparison for PCA with 99% retained variance. In Fig. 10a, CFS performs worse than combined classifiers for all training set sizes used in the experiments except for 10 and 20 where the overtraining at scale $S_2$ deteriorates the results of the combined classifier because regularization is not applied in our experiments at scales $S_2$ and $S_3$. When sufficient components are used, CFS asymptotically performs the same or even better than combined classifiers as shown in Fig. 10b.

*6.2. Combined Classifier versus CFS Using Multiresolution Histograms*

As mentioned in Sections 1 and 2, the most common approach for the construction of DPR from the feature subsets is their fusion. Due to the high dimensionality of the constructed DPR, a severe feature reduction is usually deployed in these techniques.

We compare here our proposed technique based on combined classifiers with *multiresolution histograms* (MH) as proposed by van Ginneken et al. [10]. The same as our approach, they use scale space theory and the N-jet of derivatives to construct the multiscale textures. Hence, the multiresolution technique is the same in both approaches. The difference is in the classification system as in our approach the feature subsets, which are the pixels in the patches extracted from multiscale texture after applying PCA, are directly submitted to the BCs, whereas in their approach, after extraction of the patches from multiscale textures, four histogram moments are calculated at each scale: mean, standard deviation, skewness, and kurtosis. The obtained feature subsets are fused to construct a CFS and given to a 1-NN classifier.

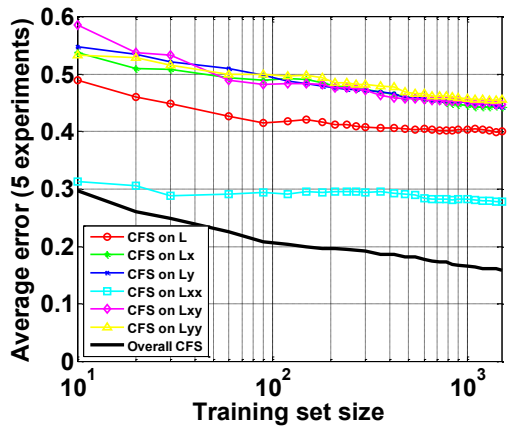

Figure 11: The learning curves for the MH approach applied to the 4-class problem of Brodatz.

To make the two approaches comparable, the construction of multiscale textures has been done in the same manner as explained in the experiment section and patches of the same size are extracted in both approaches. As in the MH approach the CFS is applied to a 1-NN classifier, the features are normalized. This is not crucial in our approach as QDC is invariant to scaling [51].

The results for the MH approach on the 4-class problem of Brodatz (Fig. 4a) are shown in Fig. 11 that can be compared with the graphs in Fig. 8a, which are the results for the proposed approach in this paper for the same textures. The superiority of the proposed approach is obvious from this comparison.

At small training set sizes, MH performs better and this is due to few features in this approach. However, the error is still too big and this means the training of the system is not yet adequate and more data samples are needed to achieve a reasonable performance.

It is emphasized here that comparison of different techniques by learning curves reveals the relative performance of two approaches for both small and large training set sizes. This is usually missing in multiresolution approaches in the literature where only comparison for a specific training set size (usually large) is reported.

6.3. Combined Classifiers versus CFS using SVMs

Kernel-based classifiers like SVMs are considered as the state-of-the-art classifier in pattern recognition. These classifiers particularly perform well in a high dimensional feature space (refer to [9, 24] for their application in texture classification). Hence, they can potentially perform well in multiresolution approaches where the feature subsets from different resolutions are concatenated. The crucial issue in using SVMs is finding a suitable kernel and the optimum trade-off parameter $C$. Radial basis function (RBF) is recommended in [52] as the first choice for the kernel. The optimum kernel width (here $\gamma$) and $C$ can be found in a particular problem by grid search [52].

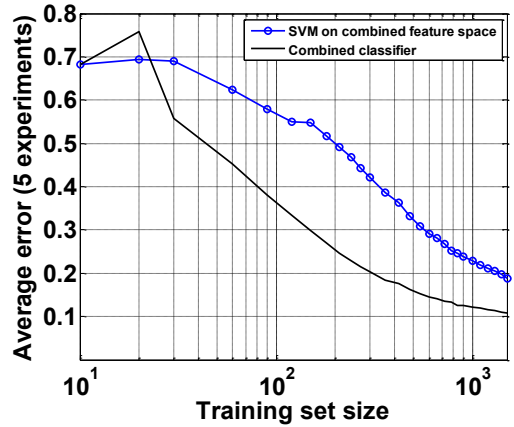

Figure 12: The comparison between combined classifier and combined feature space using SVMs in multiscale zeroth order derivative on 4-class problem of Brodatz (Fig. 4a).

To compare our technique with SVMs in multiscale, we limited our test to multiscale in the zero[th] order derivative. The construction of feature subsets is performed in the same way in both approaches, i.e., by computation of multiscale textures in the zero[th] order derivative, patch extraction from different scales, and applying PCA for feature extraction. However, the features obtained are normalized before submitting to SVMs while in our approach normalization of features is not needed.

Fig. 12 displays the comparison between the performance of the SVMs on CFS and combined classifiers for the 4-class problem of Brodatz (Fig. 4a). For the SVM, the RBF kernel width and $C$ are optimized by grid search. Since grid search is expensive, these optimal values are only calculated for four training set sizes, i.e., 10, 60, 500, and 1500. The same settings have been used in intervals around these sizes. The superiority of the combined classifier approach to SVMs is clear from Fig. 12.

## 7. Discussions and Conclusions

A two-stage multiple classifier system is proposed for the classification of small texture patches. Linear scale space theory used in this paper constructs the multiscale textures using linear operators, i.e., convolution and differentiation as given in (4). Thus, from pattern recognition point of view, whatever can be done in multiscale space can be also performed in the original space. However, additional information provided from other scales/derivatives helps to train the classifiers faster and hence using smaller training set sizes. The combined classifiers in the multiscale context with the structures proposed in this paper can further help fast training of the classifiers by submitting the feature subsets obtained at each scale directly to BCs. This is in contrast to common trend in the literature that the feature subsets from different scales are fused first and then submitted to a classifier. By using combined classifiers, since the feature subsets can be presented directly to the BCs, the training of each BC can be done using fewer data samples as the dimensionality of feature space is less than fused feature space.

On the other hand, to overcome the problems of high dimensionalities of the fused feature space in multiscale approaches, severe feature reductions are usually applied in the literature to feature subsets of every scale, e.g. by calculating histogram moments or energy. This causes that significant information is lost from the original feature subset. The performance of the new feature subset is thereby highly application dependent. Using combined classifiers again lifts this requirement for severe feature reduction as the feature subsets are not fused.

A general setup is considered in this research as our focus is on the design of a classifier/proposing a structure that can handle high dimensional feature space generated from multiresolution techniques in texture classification. Being invariant to lighting conditions or other variations is usually at the level of the design of features not classifiers. Our structure can be used with any multiresolution technique, which is designed for such variations. Also, the proposed structures can be used to combine different feature methods instead of combining the features obtained at multiscales. This can be further investigated in future research.

The proposed two-stage combined classifier provides:

1. a versatile means for grouping feature subsets in a more sensible way, e.g., different scales of the same derivative or different derivatives of the same scale.
2. a versatile means to investigate the significance of different scales/derivatives in overall performance of the multiscale classification system.
3. more flexibility because the best pair of combiners for the two levels can be chosen, as explained in Section 5.3, that can perform better than one single combiner and the best combination of feature subsets can be investigated and selected.

It is also shown that the proposed approach can perform better than SVMs in multiscale texture classification.